\documentclass[journal,onecolumn]{IEEEtran}
\ifCLASSINFOpdf
\else
\fi
\usepackage{epsfig}
\usepackage{graphicx}
\usepackage{amsmath}
\usepackage{amssymb}
\usepackage{commath}
\usepackage{bbding}
\usepackage{cite}
\usepackage{cleveref}
\usepackage{booktabs,multirow}

\usepackage{epstopdf}

\usepackage{amsmath,graphicx}
\usepackage{amsfonts}
\usepackage{subfigure}

\begin{document}
% \title{Feature-Independent Projective Image Registration in the Haar Wavelet Domain}
% \title{Projective Image Registration in the Haar Wavelet Domain Without Feature Detection and Matching}
% \title{Feature-Independent Projective Image Registration in the Haar Domain}
%\title{Projective Image Registration in the Haar Domain Without Tracking Features}
% \title{Projective Image Registration in the Haar Wavelet Domain without Corresponding Features}
%\title{Direct Projective Image Registration in the \\Haar Wavelet Domain}
% \title{Projective Image Registration Without Point Correspondences in the Haar Domain}
%--------------------------------------------------------------------------------------------- 
%--------------------------------------------------------------------------------------------- 
%--------------------------------------------------------------------------------------------- 
\title{View-Invariant Template Matching Using Homography Constraints}

\author{Sina Lotfian and Hassan Foroosh
        % <-this % stops a space
\thanks{Sina Lotfian and Hassan Foroosh are with the Department of Computer Science, University of Central Florida, Orlando,
FL, 32816 USA (e-mail: slotfian@Knights.ucf.edu, foroosh@cs.ucf.edu).}% <-this % stops a space
}

\maketitle

\begin{abstract}
Change in viewpoint is one of the major factors for variation
in object appearance across different images. Thus, view-invariant object recognition is a challenging and important image understanding task. In this paper, we propose a method that can match objects in images taken under different viewpoints. Unlike most methods in the literature, no restriction on camera orientations or internal camera parameters are imposed and no prior knowledge of 3D structure of the object is required. We prove that when two cameras take pictures of the same object from two different viewing angels, the relationship  between every quadruple of points reduces to the special case of homography with two equal eigenvalues. Based on this property, we formulate the problem as an error function that indicates how likely two sets of 2D points are projections of the same set of 3D points under two different cameras. Comprehensive set of experiments were conducted to prove the robustness of the method to noise, and evaluate its performance on real-world applications, such as face and object recognition.
\end{abstract}



\section{Introduction}
\label{sec:intro}

Object recognition from raw images given one or more examples as template(s) is a challenging problem that has important applications in diverse areas of computer vision such as image annotation \cite{Tariq_etal_2017_2,Tariq_etal_2017,tariq2013exploiting,tariq2015feature,tariq2014scene,tariq2015t}, self-localization \cite{Junejo_etal_2010,Junejo_Foroosh_2010,Junejo_Foroosh_solar2008,Junejo_Foroosh_GPS2008,junejo2006calibrating,junejo2008gps}, surveillance \cite{Junejo_etal_2007,Junejo_Foroosh_2008,Sun_etal_2012,junejo2007trajectory,sun2011motion,Ashraf_etal2012,sun2014feature,Junejo_Foroosh2007-1,Junejo_Foroosh2007-2,Junejo_Foroosh2007-3,Junejo_Foroosh2006-1,Junejo_Foroosh2006-2,ashraf2012motion,ashraf2015motion,sun2014should}, human action and interaction recognition \cite{Shen_Foroosh_2009,Ashraf_etal_2014,Ashraf_etal_2013,Sun_etal_2015,shen2008view,sun2011action,ashraf2014view,shen2008action,shen2008view-2,ashraf2013view,ashraf2010view,boyraz122014action,Shen_Foroosh_FR2008,Shen_Foroosh_pose2008,ashraf2012human}, target localization and tracking \cite{Shu_etal_2016,Milikan_etal_2017,Millikan_etal2015,shekarforoush2000multi,millikan2015initialized}, shape modeling and pattern recognition \cite{Cakmakci_etal_2008,Cakmakci_etal_2008_2,Zhang_etal_2015,Lotfian_Foroosh_2017,Morley_Foroosh2017,Ali-Foroosh2016,Ali-Foroosh2015,Einsele_Foroosh_2015,ali2016character,Cakmakci_etal2008,damkjer2014mesh},  and image-based rendering \cite{Cao_etal_2005,Cao_etal_2009,shen2006video,balci2006real,xiao20063d,moore2008learning,alnasser2006image,Alnasser_Foroosh_rend2006,fu2004expression,balci2006image,xiao2006new,cao2006synthesizing}. The problem is often exacerbated by issues such as image quality, noise, and drastic appearance changes caused by viewpoint variations. Although, preprocessing steps such as image enhancement \cite{Foroosh_2000,Foroosh_Chellappa_1999,Foroosh_etal_1996,Cao_etal_2015,berthod1994reconstruction,shekarforoush19953d,lorette1997super,shekarforoush1998multi,shekarforoush1996super,shekarforoush1995sub,shekarforoush1999conditioning,shekarforoush1998adaptive,berthod1994refining,shekarforoush1998denoising,bhutta2006blind,jain2008super,shekarforoush2000noise,shekarforoush1999super,shekarforoush1998blind},  and registration  \cite{Foroosh_etal_2002,Foroosh_2005,Balci_Foroosh_2006,Balci_Foroosh_2006_2,Alnasser_Foroosh_2008,Atalay_Foroosh_2017,Atalay_Foroosh_2017-2,shekarforoush1996subpixel,foroosh2004sub,shekarforoush1995subpixel,balci2005inferring,balci2005estimating,foroosh2003motion,Balci_Foroosh_phase2005,Foroosh_Balci_2004,foroosh2001closed,shekarforoush2000multifractal,balci2006subpixel,balci2006alignment,foroosh2004adaptive,foroosh2003adaptive}
may help in tackling some of the challenges, viewpoint variations remain by and large challenging.   

The variation in pose and viewpoint can cause  distortion in the feature space to the extent that many recognition algorithms may fail to recognize objects. The relationship between the rotation and translation of an object in the 3D world and the changes in the coordinates of pixels in the 2D image plane is also not trivial. Algorithms dealing with variation in viewpoint usually make assumptions either about change in feature space caused by relative 3D transformations, or about the position and orientation of the camera or requiring autocalibration to estimate the camera parameters \cite{Cao_Foroosh_2007,Cao_Foroosh_2006,Cao_etal_2006,Junejo_etal_2011,cao2004camera,cao2004simple,caometrology,junejo2006dissecting,junejo2007robust,cao2006self,foroosh2005self,junejo2006robust,Junejo_Foroosh_calib2008,Junejo_Foroosh_PTZ2008,Junejo_Foroosh_SolCalib2008,Ashraf_Foroosh_2008,Junejo_Foroosh_Givens2008,Lu_Foroosh2006,Balci_Foroosh_metro2005,Cao_Foroosh_calib2004,Cao_Foroosh_calib2004,cao2006camera}
from the images in order to account for viewpoint and camera parameter changes. Learning viewpoint manifolds \cite{elgammal2004inferring} \cite{bakry2014untangling} and the latent spaces for viewpoints \cite{CCA} \cite{sharma2012generalized} are two popular approaches taken by researchers for this problem, but they require simplifying assumptions in order to solve the problem. In this paper, a geometric approach is taken to address this problem and a solution in the most general case is provided.

We propose a template matching method based on image-domain relations in the projective space that can match objects across any pair of poses as long as the template image and the probe image have enough overlap for keypoint extraction. We prove that for one object seen by two cameras, with arbitrary intrinsic and extrinsic camera parameters, a restriction applies on the eigenvalues of the homography matrices associated with any quadruple of keypoint correspondences. By exploiting this constraint, an error function is introduced that is able to estimate how likely the provided reference and test images belong to the same object under different viewpoints.

The novelty of the paper can be summarized as follow:

\begin{itemize}
\renewcommand{\labelitemi}{$\bullet$}
\item We propose a template matching method that can match the given template with any inquiry image even under a wide baseline and viewpoint changes, as long as they have overlaps.

\item Unlike learning-based methods, the proposed approach does not need separate training data for each viewpoint. We also do not make any assumptions on the orientation of the cameras or their intrinsic matrices.

\end{itemize}

%\begin{figure}[t]
%\begin{center}
%  %\centering
%  \includegraphics[width=\linewidth]{Steps.png}
%  \end{center}
%
%%  \vspace{2.0cm}
%  \caption {The main steps of proposed algorithm for view-invariant template matching between the  reference and query images.}
%\label{fig:gist}
%
%\end{figure}

\section{Related Work}
\label{sec:related}

One common approach to tackle the varience in pose is to find latent spaces where the correlation between two views are maximized. Canonical correlation analysis(CCA) \cite{CCA} projects the data from two views into two low dimensional subspace which are highly correlated. Sharma et al. \cite{sharma2012generalized} have extended CCA method so that it exploits the labels of training data to find a more discriminative projection direction. Both of the mentioned methods can exploit kernels to model non-linearity. Although methods based on latent spaces have proven to be powerful tools for both multi-view image classification and multi-modal data classification, they require learning a projection direction for every viewpoint and their ability to generalize to unseen viewpoints is limited.

Another set of solutions try to fit the given 2D images to predefined 3D shapes of objects (e.g. a face) from a single view image \cite{yi2013towards}\cite{li2009maximizing}. In \cite{asthana2011fully} authors propose a 3D pose normalization for face recognition in order to make it robust to variation in pose. \cite{schels2012learning} exploits 3D CAD models to detect and find the pose of objects such as bikes and cars. The use of these methods are restricted to objects with available predetermined 3D models. A rather interesting solution was proposed by \cite{castillo2009using} that does not require 3D reconstruction of the face, instead they use the cost of stereo matching as their error function. However, they make the assumption that epipolar lines are horizontal which does not hold true for the object recognition in the general case.

Ideally, we are in search of view-invariant recognition algorithms that require few training data (hopefully one shot learning) \textbf{(OSL)}, generalizable to unseen view-points \textbf{(GUV)}, work on objects without known 3D structure, \textbf{(3DFree)} and invariant to the internal camera parameters \textbf{(IICP)}. Table \ref{table:t1} compares the various classes of algorithms described above, in terms of these desired properties.

In this paper, we take a geometric approach to the problem of viewpoint variation. Our work is inspired by Shen et al.\cite{shen2008view}, who used homographgy constraints to recognize body pose transitions between two successive frames of two video cameras, observing human actions. Although, we are not dealing with video frames in this work, we show that the concept can be extended also to a pair of still images of a rigid object (i.e. instead of dealing with moving points in space viewed by two pairs of frames (4 images), we can extend the idea to recognizing a rigid object from two images. The key to achieve this extension is to consider quadruple of points in each camera image, instead of triplets of points in two frames of each camera The result is a rigid object recognition method that can handle unknown viewpoints and internal camera parameters. 

\begin{table}[]
\centering
\caption{ Desirable properties of some view-invariant recognition algorithms. If an algrothims satisfy the condition it is indicated by a check-mark. }
\label{table:t1}
\begin{tabular}{|l|c|c|c|c|}
\hline
     & \multicolumn{1}{l|}{GUV} & \multicolumn{1}{l|}{OSL} & \multicolumn{1}{l|}{3DFREE} & \multicolumn{1}{l|}{IICP} \\ \hline
CCA \cite{CCA}  &                          &                          & \checkmark                           &                           \\ \hline
GMA \cite{sharma2012generalized} &                          &                          & \checkmark                            &                           \\ \hline
DPFD \cite{sanyal2015discriminative}& \checkmark                         &                          & \checkmark                            &                           \\ \hline
Castillo et al \cite{castillo2009using} & \checkmark                         & \checkmark                         & \checkmark                            &                           \\ \hline
Schels et al \cite{schels2012learning}& \checkmark                         &   \checkmark                        &                             &                           \\ \hline
Ours & \checkmark                         &  \checkmark                         &                            \checkmark  &     \checkmark                       \\ \hline
\end{tabular}
\end{table}

%\begin{figure*}[t!]
%  \centering
%  \includegraphics[keepaspectratio, width=0.6\textwidth]{./figure1.png}
%  \caption{Quadruple of points being viewed by two different cameras. In the part (a) you can see the plane $l_1$ is used to map triplet of points between two image using the homography $H_1$. The same rule is applied in part (b) for plane $l_2$ and homography $H_2$. Note that the red line at the intersection of two planes as well as epipoles are fixed under  $H = H_1H_{2}^{-1}$. Hence, it is a planar homology with two equal eigenvalues.}
%  \label{fig:Planes}
%\end{figure*}

\begin{figure*}
\centering     %%% not \center
\subfigure[]{\label{subfig:planes_a}\includegraphics[width=50mm]{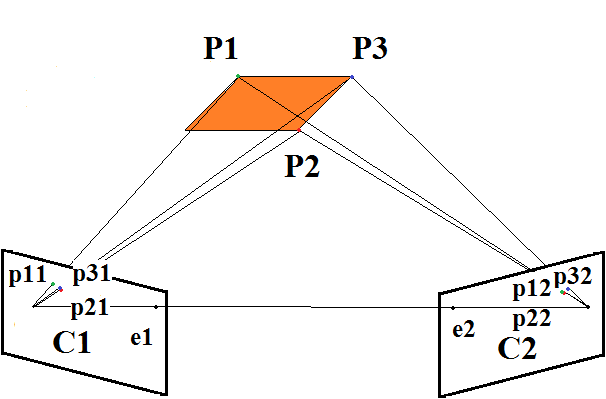}} \hspace{0.5mm}
\subfigure[]{\label{subfig:planes_b}\includegraphics[width=50mm]{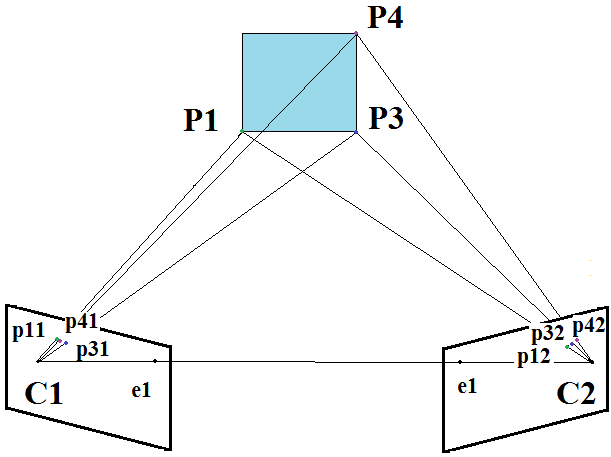}} \hspace{0.5mm}
\subfigure[]{\label{subfig:planes_c}\includegraphics[width=50mm]{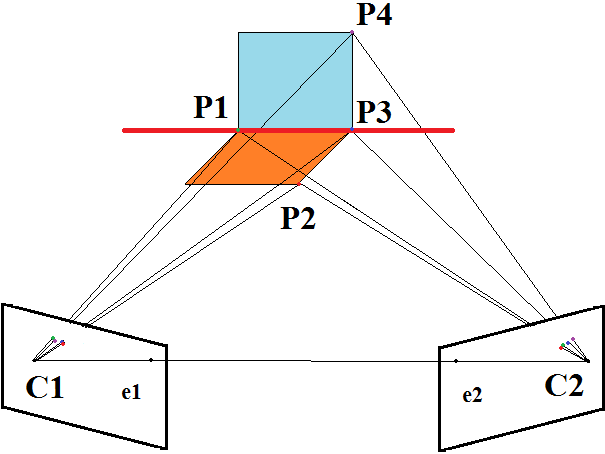}}
\caption{Quadruple of points being viewed by two different cameras. In the part (a) you can see the plane $l_1$ is used to map triplet of points between two image using the homography $H_1$. The same rule is applied in part (b) for plane $l_2$ and homography $H_2$. Note that the red line at the intersection of two planes as well as epipoles are fixed under  $H = H_1H_{2}^{-1}$. Hence, it is a planar homology with two equal eigenvalues.}
\label{fig:Planes}
\end{figure*}

\section{Proposed Method}
\label{sec:proposed}

Given a reference image ($I_{r}$)  and a query image ($I_{q}$), our goal is to determine if they belong to the same 3D object under two different viewpoints or not. First, point correspondences are extracted between $I_{r}$  and $I_{q}$, and represented as $S= \{ ({\bf p}_{r1},{\bf p}_{q1}), ({\bf p}_{r2},{\bf p}_{q2}), ... ({\bf p}_{rn}, {\bf p}_{rn})\}$. Such correspondences can be obtained from any keypoint extraction and matching algorithm such as SIFT\cite{lowe2004distinctive}, SURF\cite{bay2006surf} or Harris\cite{harris1988combined}.  For more clarity, we use upper case letters for 3D coordinates and lower case for 2D coordinates on the image plane. We introduce an error function that in the ideal case vanishes, when there exist a unique 3D configuration of points which map to the extracted 2D keypoint correspondences. Conversely, the value of the error function increases, if such 3D configuration is not possible. Furthermore, the proposed error function is fully projective (i.e. fully defined in the image domain) and hence is invariant to camera positions and its internal parameters.

Consider the object shown in Figure \ref{fig:Planes}, which consist of four 3D points $\{{\bf P}_1,{\bf P}_2,{\bf P}_3,{\bf P}_4\}$ in general positions. Two cameras (C1 and C2) that are located in two different coordinates are imaging this object as $I_r$ and $I_q$. In the most general case, the two cameras would be projective with 11 degrees of freedom (i.e. different intrinsic parameters and arbitrary orientations in the 3D space). Two key observations that  lead to the proposed solution are: (1) Any three of the quadruple of points define a plane in the 3D space that induces a homography between the two cameras; (2) With a quadruple of points one can obtain 4 such planes, i.e. two pairs of homographies. Each pair of homographies plays a similar role as a moving plane considered in \cite{shen2008view}, except that in our case instead of a single plane moving in time, we are considering the dual case of two planes in a rigid body. Since this case is dual to the problem considered by \cite{shen2008view}, the construct remains the same. We illustrate this using the example of Figure \ref{fig:Planes}.

Let two planes $\pi_1$ (orange) in Figure \ref{subfig:planes_a} and $\pi_2$ (blue) in Figure \ref{subfig:planes_b} correspond to the triplets of 3D points $\{{\bf P}_1,{\bf P}_2,{\bf P}_3\}$ and $\{{\bf P}_1, {\bf P}_2,{\bf P}_4\}$, respectively. Let the corresponding image points be $p_r = \{{\bf p}_{r1},{\bf p}_{r2},{\bf p}_{r3},{\bf p}_{r4}\}$ and $p_q = \{{\bf p}_{q1},{\bf p}_{q2},{\bf p}_{q3},{\bf p}_{q4}\}$. We assume that no three projected points are co-linear in either views. Let also ${\bf e}_1$ and ${\bf e}_2$ denote the epipoles in the two images. Since epipoles are mapped across two images by the homography induced by any plane in the scene, we have
\begin{eqnarray}
{\bf H}_1 {\bf p}_{ri}&=&{\bf p}_{qi}, \;\;\; i=1,2,3\\
{\bf H}_1 {\bf e}_1 &=& {\bf e}_2
\end{eqnarray}  
\begin{eqnarray}
{\bf H}_2 {\bf p}_{ri}&=&{\bf p}_{qi}, \;\;\; i=1,2,4\\
{\bf H}_2 {\bf e}e_1 &=& {\bf e}_2
\end{eqnarray}  
These equations yield a pair of homographies through which we can define ${\cal H} = {\bf H}_1 {\bf H}_{2}^{-1}$. \\

{\bf Proposition 1}: ${\cal H}$ will reduce to a homology if and only if the presumed point correspondences $p_r$ and $p_q$ are images of the same 3D point configuration.\\

The immediate consequence of this observation is that two of the eigenvalues of ${\cal H}$ must be equal if the presumed point correspondences $p_r$ and $p_q$ are images of the same 3D point configuration. This allows us to define a cost function that would  make it possible to determine if a set of image points and their matching correspondences from a template image are originated from the same 3D object. Suppose we have $m$ such template images  and we establish $n$ putative point correspondences between the query image and each reference template. One can then define $K=\dbinom{n}{4}$ quadruples of point correspondences, yielding a total of $2K$ matrices, ${\cal H}_{km}$, $k=1,...,2K$, for each template $m \in \{1,...,M\}$. Let $\epsilon_1({\cal H}_{km})$ and $\epsilon_2({\cal H}_{km})$ be the two closest eigenvalues of the matrix ${\cal H}_{km}$. Finding the optimal matching template $\hat m$ is then a labeling process that would be given by:

%\begin{equation}
%E({\cal H}_i)=\sum_1^{2k} \frac{|\epsilon_1({\cal H}_i)-\epsilon_2({\cal H}_i)|}{|\epsilon_1({\cal H}_i)+\epsilon_2({\cal H}_i)|}
%\end{equation}

\begin{equation}
\hat{m}=\arg\min_{m\in\{1,...,M\}} \sum_{k=1}^{2K} \frac{|\epsilon_1({\cal H}_{km})-\epsilon_2({\cal H}_{km})|}{|\epsilon_1({\cal H}_{km})+\epsilon_2({\cal H}_{km})|} \label{eq:xdef2}
\end{equation}

\section{Experimental Results}
\label{sec:Experimental}

In this section, the performance of the proposed method on both synthetic and real-world datasets is demonstrated with wide applications such as object and face recognition.

\begin{figure*}
\centering     %%% not \center
\subfigure[$\sigma = 0$]{\label{fig:a}\includegraphics[width=50mm]{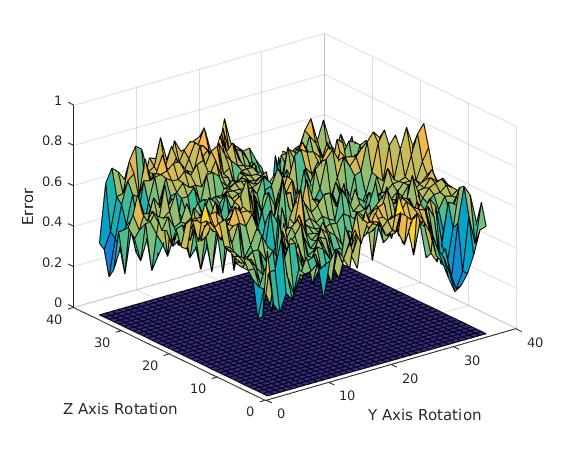}}
\subfigure[$\sigma = 6$]{\label{fig:b}\includegraphics[width=50mm]{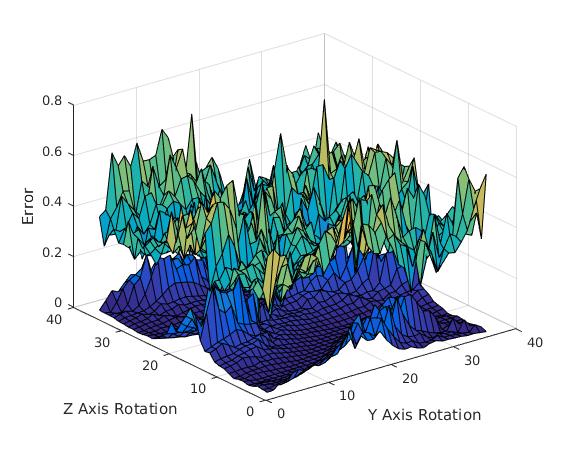}}
\subfigure[$\sigma = 12$]{\label{fig:c}\includegraphics[width=50mm]{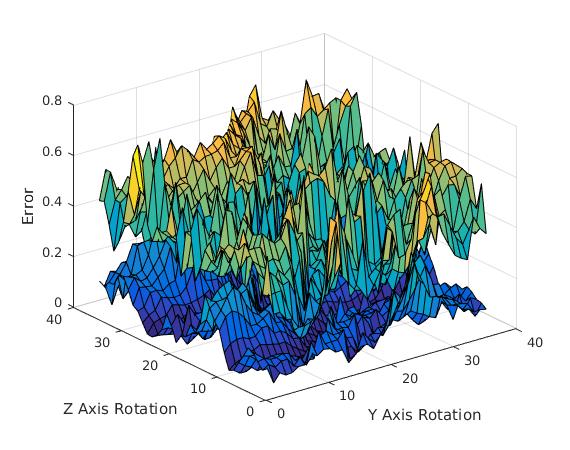}}
\caption{The error surface for different viewing angles for a) $\sigma = 0$, b) $\sigma = 6$, c) $\sigma = 12$. Without the presence of Gaussian noise the error function would vanish for same 3D point configurations between the query points and the template points. As the Gaussian noise is added to the points in the image frame, the error grows, creating a possible source of uncertainty. }
\label{fig:error}
\end{figure*}

\subsection{Synthetic Data}

In order to understand the behavior of the error function in equation \ref{eq:xdef2} in the presence of noise in key point localization, the process of projection of 3D points on the image plane is simulated using the pinhole camera model. The point clouds used for generating the synthetic objects are obtained from the BigBIRD \cite{singh2014bigbird} dataset, which consist of RGBD images of objects sampled on the the viewing hemisphere. Object 'Advil' is chosen as the positive example and the object 'Syrup' is chosen as negative example. It is expected that the error measure for 'Advil-Advil' pair will be lower that 'Advil-Syrup' pair.

Two cameras are used to generate synthetic images on the image plane. The first camera which is the reference camera is fixed at the world origin and is looking at the Z axis. The second camera or the test camera is moving on the viewing hemisphere. This is achieved by rotating the reference camera around $Y$ and $Z$ axis. Since the number of points in the cloud is over one thousand, we randomly choose 8 points as the keypoints and project only these 8 points on the image plane. The focal lengths for both cameras change randomly in the range $1000 \pm 100$. Then by adding Gaussian noise to the position of keypoints on the image plane, we measure the robustness of the algorithm. 

In figure \ref{fig:error} the matching score for different viewing angles are plotted for both the matching query-template pair (the surface below) and the non-matching query-template pair (surface above). It can be observed that for the matching pair the error is almost zero, while for non-matching pairs the error is high. To find out the extent of separation between these (i.e. ability to distinguish between a correct and incorrect match), we added Gaussian noise to the position of the keypoints in the image planes. It can be observed that as the noise variance increases, the two error surfaces get closer and the ditinction between true match and a false match becomes harder. Our experiments show that we can handle noise strength of up to about $\sigma = 12$ which roughly equates the correspondences being 24 pixels off.

\subsection{Real-World Data}

We also tested the proposed method on real datasets, including a 3D multi-view object recognition dataset, and two multi-view face recognition datasets. The first dataset is coil-20 \cite{nene1996columbia}, which consists of 20 classes, each taken with the object rotated 5 degree on a turn-table. Pointing04 \cite{gourier2004estimating} and UMIST\cite{graham1998characterising} are two multi-view face dataset used to evaluate the proposed method. UMIST consists of 575 faces from 20 different persons taken under different conditions. Pointing04 contains 2690 face photos taken from 15 people. The Pointing04 face rotation has more degrees of freedom and images are taken with and without glasses. Keypoints are extracted using the popular SIFT \cite{lowe2004distinctive} descriptor and matched using a nearest neighbor method. Note that unlike many methods in the literature the proposed algorithm does not assume that the coordinates of facial keypoints, such as nose and lips are given, and it only relies on the features extracted by SIFT, which may lay anywhere on the face.

Although in theory our method needs only one template per class to match two images, there has to be enough overlap between the query and reference image so that the keypoint extractor algorithm can find enough mutual keypoints in both images. Therefore, in each dataset for every class, 8 images are chosen as the templates and the rest are used as query images. For instance, in the Coil-20 dataset 8 images are taken as templates and 64 used for the test phase. The overall accuracy for all dataset are provided in \ref{table:accuracy}. 

\begin{table}[]
\centering
\caption{Object and face recognition dataset and their associated accuracy and size}
\label{table:accuracy}
\begin{tabular}{|l|l|l|l|}
\hline
\multicolumn{1}{|c|}{Dataset} & \multicolumn{1}{c|}{Accuracy} & \multicolumn{1}{c|}{No. of Classes} & \multicolumn{1}{c|}{Dataset Size} \\ \hline
Coil-20                       & 86.1                          & 20                                  & 1440                              \\ \hline
Pointing04                      & 77.7                          & 15                                  & 2690                              \\ \hline
UMIST                         & 92.5                          & 20                                  & 565                               \\ \hline
\end{tabular}
\end{table}

\section{Conclusion}

In this paper, a new view-invariant template-matching method is introduced that imposes no restrictions on external or internal camera parameters. The robustness of the algorithm has been tested by adding Gaussian noise to the coordinates of the keypoints on the image to simulate the behavior of error in keypoint localization. Finally, the accuracy of the method on object and face recognition was tested, producing remarkable good results.

%--------------------------------------------------------------------------------------------- 

{%% \small
\bibliographystyle{plain}
\bibliography{strings,refs,foroosh}
}

\end{document}